\documentclass[lettersize,journal]{IEEEtran}
\usepackage{amsmath,amsfonts}
\usepackage{algorithm}
\usepackage{array}
\usepackage[caption=false,font=normalsize,labelfont=sf,textfont=sf]{subfig}
\usepackage{textcomp}
\usepackage{stfloats}
\usepackage{url}
\usepackage{verbatim}
\usepackage{graphicx}
\usepackage{cite}
\usepackage{xspace}
\usepackage{xcolor}
\usepackage{hyperref}
\usepackage{algorithm}
\usepackage{amsmath,amssymb,amsfonts}
\usepackage{algpseudocode}  
\usepackage{booktabs}
\usepackage{array,multirow}
\usepackage{arydshln}
\usepackage{bm}

\newcommand{\attacker}{\textsc{CAAP}} 

\newcommand{\codeurl}{\url{https://github.com/ryliu68/CAAP}}

\begin{document}

\title{\attacker: Capture-Aware Adversarial Patch Attacks on Palmprint Recognition Models}


\author{Renyang Liu,
        Jiale Li,
        Jie Zhang, 
        Cong Wu,
        Xiaojun Jia,
        Shuxin Li,
        Wei Zhou,~\IEEEmembership{Member,~IEEE,}        
        Kwok-Yan Lam,~\IEEEmembership{Senior Member,~IEEE,}
        See-kiong Ng,~\IEEEmembership{Member,~IEEE}

\IEEEcompsocitemizethanks{
\IEEEcompsocthanksitem R. Liu, J. Li, and S.K. Ng are with the Institute of Data Science, National University of Singapore, Singapore 117602, Singapore (e-mail: \{ryliu,seekiong\}@nus.edu.sg, e1351034@u.nus.edu)
\IEEEcompsocthanksitem J. Zhang is with the Centre for Frontier AI Research (CFAR), A*STAR, Singapore, 138634 (e-mail: zhang\_jie@cfar.a-star.edu.sg).
\IEEEcompsocthanksitem C. Wu is with the School of Cyber Science and Engineering, Wuhan University, China, 430072 (e-mail: cnacwu@whu.edu.cn).
\IEEEcompsocthanksitem X. Jia, S. Li, and K.Y. Lam are with the College of Computing and Data Science, Nanyang Technological University, Singapore, 639798 (e-mail: jiaxiaojunqaq@gmail.com, \{shuxin001,kwokyan.lam\}@ntu.edu.sg).
\IEEEcompsocthanksitem W. Zhou is with the School of Engineering, Yunnan University, Kunming 650500, China (e-mail: zwei@ynu.edu.cn).
}
}

\maketitle

\begin{abstract}
Palmprint recognition is increasingly deployed in security-critical applications, such as access control and palm-based payment, due to its contactless acquisition and highly discriminative ridge-and-crease textures. However, the robustness of deep palmprint recognition systems against physically realizable attacks remains insufficiently understood. Existing studies are largely confined to the digital setting and do not adequately account for two practical factors: the texture-dominant nature of palmprint recognition and the capture-induced distortions introduced during physical acquisition.
To address this gap, we propose \attacker, a capture-aware adversarial patch framework for palmprint recognition. \attacker\ learns a universal patch that can be reused across inputs while remaining effective under realistic acquisition variation. To better accommodate the structural characteristics of palmprints, the framework adopts a cross-shaped patch topology, which enlarges spatial coverage under a fixed pixel budget and more effectively disrupts long-range texture continuity. \attacker\ further integrates three modules: an Adaptive Spatial Transformer (ASIT) for input-conditioned patch rendering, a Radiometric Synthesis module (RaS) for stochastic capture-aware simulation, and a Multi-Scale Dual-Invariant Feature Extractor (MS-DIFE) for feature-level identity-disruptive guidance.
We evaluate \attacker\ on two public datasets, Tongji and IITD, and an in-house dataset, AISEC, against both generic CNN backbones and palmprint-specific recognition models. Extensive experiments show that \attacker\ achieves strong untargeted and targeted attack performance together with favorable cross-model and cross-dataset transferability. The results further show that, although adversarial training can partially reduce the attack success rate, substantial residual vulnerability remains. These findings indicate that deep palmprint recognition systems remain vulnerable to physically realizable, capture-aware adversarial patch attacks, underscoring the need for more effective defenses in practical deployment.
Code available at \textcolor{blue}{\codeurl}.
\end{abstract}


\section{Introduction}
\label{sec:intro}
\IEEEPARstart{W}{ith} the rapid deployment of biometric authentication in security-critical applications, recognition systems based on faces, fingerprints, irises, and palms have become integral to modern access-control and payment infrastructures~\cite{TBIOM_20/contactless}. Among these modalities, palmprint recognition has attracted increasing attention because it supports contactless acquisition, offers relatively high user acceptance, and provides rich ridge-and-crease textures for identity discrimination~\cite{TBIOM_20/contactless,PR_09/Kong_survey}. Recent deployments further indicate its practical viability at scale, spanning commercial payment and access-control scenarios around the world~\cite{idtechwire2025_alipay_pl1,visa2024_tencent_sg,investopedia2023_amazonone_wholefoods,zaobao2019_octobox}. This growing adoption elevates the security stakes because biometric traits are inherently non-revocable and the consequences of compromise are difficult to mitigate once a system is systematically targeted.

Driven by this demand, palmprint recognition relies on pattern-recognition pipelines and deep feature extractors~\cite{PR_09/Kong_survey,TBIOM_20/contactless,tifs_19/PalmNet,tifs_23/ccnet,tim_23/CO3Net}. Although deep models improve recognition accuracy under benign capture variations~\cite{TBIOM_20/contactless,tifs_19/PalmNet,tifs_23/ccnet,tim_23/CO3Net}, adversarial machine learning has shown that deep recognition systems can be manipulated by carefully crafted inputs, leading to misclassification or authentication failures~\cite{TIFS_24/CDMA,iclr_18/PGD}. In biometric settings, such vulnerabilities are particularly concerning because they directly weaken access-control guarantees and their impact is amplified by the non-replaceable nature of biometric identifiers.

Despite these concerns, robustness evaluation for deep palmprint recognition remains comparatively limited, especially in the physical setting~\cite{TIFS_23/MSPA,JMIS_22/presentation}. Existing studies are largely restricted to digital attacks and do not adequately consider physically realizable adversarial perturbations that must remain effective throughout the full capture pipeline~\cite{TIFS_23/MSPA,JMIS_22/presentation}. Such attacks are practically important because contactless palmprint systems are deployed in real access-control and payment scenarios, where attacks would occur through the acquisition process rather than through digital modification of ROI images. In practical deployments, these attacks are affected by printing and imaging processes, as well as by environmental factors such as hand pose, capture distance, illumination, and sensor noise. Moreover, generic patches with compact spatial support may be ill-suited to palmprint recognition. Palmprint models are strongly driven by texture cues and depend on global ridge statistics and long-range line continuity across the palm region~\cite{PR_09/Kong_survey,CVPR_05/Ordinal,PR_08/rloc}. Consequently, small block-like patches may be treated as localized artifacts and may fail to consistently disrupt the global texture representations exploited by palmprint models, especially after capture-induced geometric and photometric distortions. A further limitation is that many existing attacks are instance-specific, requiring per-image optimization to achieve high success rates. This requirement is costly in general and particularly impractical in physical settings, where the artifact must be fabricated once and reused across users and capture conditions. Therefore, evaluations limited to digital attacks~\cite{TIFS_23/MSPA} or generic patch designs~\cite{arXiv_17/AdvPatch} may systematically underestimate the real-world vulnerability of palmprint recognition systems.

To address these limitations, we propose \attacker, a capture-aware adversarial patch framework for palmprint recognition. Specifically, \attacker\ learns a universal adversarial patch that is optimized once and reused across inputs, thereby avoiding the need for instance-specific patch optimization. To better match the texture-dominant characteristics of palmprints, the framework adopts a cross-shaped patch topology under a fixed pixel budget, which enlarges spatial coverage and more effectively disrupts long-range ridge-and-line continuity. \attacker\ further integrates three components that are tailored to physically realizable attacks: ASIT performs input-conditioned patch rendering, RaS introduces stochastic capture-aware simulation during training, and MS-DIFE provides multi-scale feature guidance for identity-disruptive optimization. Together, these components form a capture-aware optimization framework for learning physically robust adversarial patches.

We evaluate \attacker\ on two public palmprint datasets, Tongji and IITD, and an in-house dataset, AISEC, across diverse victim architectures, including generic CNN backbones and palmprint-specific recognition networks. The results show that \attacker\ achieves strong untargeted and targeted attack performance together with favorable cross-model and cross-dataset transferability. They further show that, although adversarial training can partially reduce the attack success rate, substantial residual vulnerability remains. Our main contributions are summarized as follows:

\begin{itemize}
    \item We introduce \attacker, a capture-aware universal adversarial patch framework for palmprint recognition, designed for physically realizable and reusable attacks.

    \item We develop a palmprint-oriented attack design that combines a cross-shaped patch topology, input-conditioned patch rendering, stochastic capture-aware simulation, and multi-scale feature guidance to improve physical robustness and attack effectiveness.

    \item We conduct extensive experiments on public and in-house datasets, covering untargeted and targeted attacks, cross-model and cross-dataset transferability, and robustness under adversarial training, thereby providing practical insights into the physical vulnerability of deep palmprint recognition systems.
\end{itemize}

The rest of this paper is organized as follows. Section~\ref{sec:related_work} reviews related work on palmprint recognition and adversarial attacks. Section~\ref{sec:prelim} presents the preliminaries. Section~\ref{sec:method} presents \attacker, including its patch topology, ASIT, RaS, and MS-DIFE. Section~\ref{sec:evaluation} presents the experimental results and analysis. Section~\ref{sec:conclusion} concludes the paper and discusses future directions.

\section{Related Work}
\label{sec:related_work}

Palmprint recognition relies on discriminative palmar cues, including principal lines, wrinkles, ridge-level textures, and their spatial organization~\cite{PR_09/Kong_survey}. Early studies mainly follow conventional pipelines with ROI localization, hand-crafted feature extraction, and template matching, whereas more recent work has shifted toward deep representation learning for contactless and unconstrained palmprint recognition~\cite{PR_09/Kong_survey,TBIOM_20/contactless,tifs_19/PalmNet,tifs_23/ccnet,tim_23/CO3Net}.
\subsection{Palmprint Recognition}

Palmprint recognition relies on discriminative palmar cues, including principal lines, wrinkles, ridge-level textures, and their spatial organization~\cite{PR_09/Kong_survey}. Early studies mainly follow conventional pipelines with ROI localization, hand-crafted feature extraction, and template matching, whereas more recent work has shifted toward deep representation learning for contactless and unconstrained palmprint recognition~\cite{PR_09/Kong_survey,TBIOM_20/contactless,tifs_19/PalmNet,tifs_23/ccnet,tim_23/CO3Net}.

\paragraph{Traditional representations}
Traditional palmprint recognition methods emphasize robust encoding of line and texture structures for efficient matching. Representative designs include orientation- and phase-based coding schemes such as OPI~\cite{TPAMI_03/OPI}, Competitive Coding~\cite{ICPR_04/Comp_coding}, Fusion Code~\cite{PR_06/Fusion}, Ordinal Code~\cite{CVPR_05/Ordinal}, and RLOC~\cite{PR_08/rloc}. These methods collectively show that palmprint recognition depends heavily on structured line patterns and local orientation statistics, which remain core discriminative cues even in later deep-learning-based systems.

\paragraph{Deep learning-based recognition}
Deep models replace fixed hand-crafted encodings with data-driven representations that jointly capture local texture details and larger-scale palm structures. DLRF~\cite{TBBIS_20/LiuK20} learns residual embeddings for contactless palmprint identification under metric supervision. PalmNet~\cite{tifs_19/PalmNet} incorporates classical priors such as Gabor filtering and PCA-inspired dimensionality reduction into a convolutional architecture. CompNet~\cite{spl_23/CompNet}, CCNet~\cite{tifs_23/ccnet}, and CO3Net~\cite{tim_23/CO3Net} further strengthen competitive and contrastive modeling to improve discriminability under unconstrained capture conditions. In parallel, deployment-oriented studies have addressed practical issues such as efficiency, open-set generalization, cross-device robustness, and multiview modeling, as exemplified by EEPNet~\cite{PRL_22/EEPNet}, W2ML~\cite{PR_22/W2ML}, Palm-ID~\cite{TIFS_24/Palm-ID}, cross-smartphone recognition with self-paced CycleGAN~\cite{TIFS_23/CycleGAN_atten}, Semi-CPRN~\cite{TVC_23/Semi-CPRN}, and SSL\_RMPR~\cite{TNNLS_24/SSL_RMPR}. Overall, the palmprint-recognition literature has focused primarily on improving accuracy, efficiency, and generalization in practical acquisition settings.

\subsection{Adversarial Attacks on Palmprint Recognition Systems}
The vulnerability of deep neural networks to adversarial perturbations has been widely established in computer vision~\cite{iclr_18/PGD,TMM_25/STBA}. Palmprint-specific studies, however, remain relatively limited.  
MSPA~\cite{TIFS_23/MSPA} studies adversarial-example generation for multispectral palmprints within a joint attack-and-defense framework.
Cui \emph{et al.}~\cite{ICME_25/enhanced} further improve palmprint attack generation by separately considering visible and less visible identity cues. 
Related evidence also comes from presentation-attack and anti-spoofing studies in palmprint and related palm-biometric systems, which consider re-imaging attacks or liveness-related security layers and evaluate how manipulated samples degrade after re-acquisition~\cite{JMIS_22/presentation,ICIP_23/palmprint_antispoofing}.
These studies are complementary to our setting, as they address liveness or presentation-attack detection rather than the robustness of the palmprint recognition model itself.
However, these studies do not optimize physically robust adversarial patches for direct deployment against palmprint recognition systems.
A recent review of image-level attacks on palmprint recognition likewise suggests that systematic study of adversarial threats in this modality remains at an early stage~\cite{zhang2026review_palmprint_attacks}.
Taken together, existing palmprint-security studies have demonstrated vulnerability in digital and presentation-attack settings, but physically optimized and transformation-robust adversarial attacks remain underexplored.

\subsection{Physical Adversarial Perturbations and Patch Attacks}
Physical adversarial attacks seek perturbations that remain effective after real-world acquisition processes such as printing, placement, and imaging~\cite{iclrw_17/Adversarial}. Beyond input-specific perturbations, universal adversarial perturbations provide an input-agnostic threat model and motivate attacks that can generalize across samples~\cite{CVPR_17/UAP}. For robust physical optimization, expectation-over-transformation (EOT) formalizes stochastic geometric and photometric transformations during attack generation~\cite{ICML_18/Synthesizing}. A particularly important line is adversarial patch attacks, where a localized pattern is optimized to consistently fool a model when placed in the scene~\cite{arXiv_17/AdvPatch}. Subsequent studies extend this paradigm to safety-critical settings and improve its robustness, stealthiness, or realism through physical-world evaluation, perceptual constraints, generative priors, and naturalistic appearance design~\cite{CVPR_18/Robust,SafeAI_AAAI_19/DPATCH,CVPRW_19/Fooling,AAAI_19/Perceptual-Sensitive,CVPR_20/AdvCam,ICCV_21/Naturalistic}. 
More recent work also examines the effect of patch geometry itself; for example, cross-shaped patches have been shown to improve attack efficacy relative to square patches in broader vision settings~\cite{TCSVT_24/CSPA}.
In biometrics, physically realizable accessories and artifacts have also been studied for face-recognition attacks~\cite{CCS_16/Accessorize,ICPR_20/AdvHat}. However, these physical patch designs are largely developed for semantic object or face recognition rather than for palmprint recognition, where the victim model relies more heavily on structured ridge-and-line patterns than on localized semantic parts.


Overall, the prior literature shows that palmprint recognition has evolved from hand-crafted coding schemes to powerful deep models under increasingly practical acquisition settings~\cite{PR_09/Kong_survey,TBIOM_20/contactless,tifs_19/PalmNet,tifs_23/ccnet,tim_23/CO3Net}, and that palmprint-security studies have confirmed vulnerability to adversarial-example, presentation-attack, and related anti-spoofing threats~\cite{TIFS_23/MSPA,ICME_25/enhanced,JMIS_22/presentation,ICIP_23/palmprint_antispoofing,zhang2026review_palmprint_attacks}. At the same time, the broader adversarial-attack literature provides tools for physically robust and transformation-aware patch optimization~\cite{ICML_18/Synthesizing,arXiv_17/AdvPatch,TCSVT_24/CSPA}. What remains missing is a capture-aware adversarial patch framework explicitly tailored to palmprint recognition.


\section{Preliminaries}
\label{sec:prelim}
This section presents the background and formal setup for our study of adversarial patch attacks against palmprint recognition systems. We first summarize the ROI-based input convention commonly adopted in contactless palmprint recognition.
We then present a generic formulation of a universal patch attack under stochastic acquisition transformations. Finally, we specify the threat model adopted throughout the paper.

\subsection{Palmprint Recognition and ROI Convention}
A contactless palmprint recognition system typically captures an image of the hand and extracts an aligned region of interest (ROI) for subsequent recognition~\cite{PR_09/Kong_survey,TSMCS_26/gao_survey}. 
The ROI partially normalizes hand pose and translation while preserving discriminative ridge-and-crease structures~\cite{PR_09/Kong_survey}. 
Modern recognizers operate on ROI images and map each ROI to a feature embedding or class score for matching or identification~\cite{tifs_19/PalmNet,tifs_23/ccnet,TBBIS_20/LiuK20,spl_23/CompNet}. 
Throughout this paper, we use $x \in [0,1]^{H \times W}$ to denote a preprocessed ROI image. 
Unless otherwise stated, all palmprint images in this paper refer to aligned ROI images provided by the datasets or obtained through standard preprocessing, and no additional ROI extraction is performed during attack optimization or evaluation.

\subsection{Formulation of Capture-Aware Physical Patch Attacks}
We consider physically realizable patch attacks whose effect is represented in the ROI domain and must remain effective under acquisition variations such as pose change, illumination variation, and sensor noise.
Given an ROI image $x$ with identity label $y$ and a palmprint recognizer $f(\cdot)$, the attacker seeks a universal patch that is optimized once and reused across inputs.


\paragraph{Patch parameterization}
Let $P \in [0,1]^{H_p \times W_p}$ denote the learnable patch texture, and let $M \in \{0,1\}^{H_p \times W_p}$ denote a fixed binary mask that specifies the patch topology. The effective patch content is given by $P \otimes M$, where $\otimes$ denotes element-wise multiplication.

\paragraph{Physical rendering under stochastic transformations}
Let $\theta=(\theta_{\mathrm{geo}},\theta_{\mathrm{pho}})$ denote the input-conditioned rendering parameters, where $\theta_{\mathrm{geo}}$ and $\theta_{\mathrm{pho}}$ govern geometric transformation and patch-local photometric calibration, respectively.
Let $\mathcal{S}{\theta{\mathrm{geo}},\theta_{\mathrm{pho}}}(\cdot)$ denote the differentiable rendering operator that composites the transformed patch onto the ROI image, and let $\mathcal{A}_{\xi}(\cdot)$ denote the stochastic capture model parameterized by $\xi$.
A single rendered adversarial sample is defined as
\begin{equation}
x_{\mathrm{adv}}(\theta,\xi)
=
\mathcal{A}_{\xi}\!\left(\mathcal{S}_{\theta_{\mathrm{geo}},\theta_{\mathrm{pho}}}(x,P,M)\right).
\label{eq:prelim_render}
\end{equation}
Here, $\theta$ governs input-conditioned patch rendering, whereas $\xi$ captures stochastic acquisition-time variation.

\paragraph{EOT objective for a universal patch}

To obtain a patch that remains effective across inputs and acquisition conditions, we optimize $P$ under an expectation-over-transformation (EoT) formulation~\cite{ICML_18/Synthesizing}. Let $\mathcal{L}_{\mathrm{adv}}(\cdot)$ denote a generic attack loss, instantiated differently for untargeted and targeted attacks. The resulting optimization problem is
\begin{equation}
\min_{P}\ 
\mathbb{E}_{(x,y)}\mathbb{E}_{\xi}\!\left[\mathcal{L}_{\mathrm{adv}}\!\left(x_{\mathrm{adv}}(\theta(x),\xi),y;f\right)\right].
\end{equation}

The expectation over inputs formalizes universality, while the expectation over transformations enforces robustness to acquisition variation. Section~\ref{sec:method} instantiates this generic formulation with the specific modules used in \attacker.

\subsection{Threat Model}
We consider an attacker who fabricates a physical patch and places it on the palm region during image acquisition, but cannot modify the victim model, its training data, or the capture hardware.

\paragraph{Victim system}
The victim is a deep learning-based palmprint recognizer $f(\cdot)$ that operates on aligned grayscale ROI images and outputs an identity prediction for each input. Unless otherwise stated, we do not assume specialized adversarial defenses during attack generation or evaluation.

\paragraph{Adversary knowledge and capability}
We adopt a white-box optimization setting during patch optimization, including access to the victim recognizer's architecture, logits, and gradients, in order to characterize worst-case vulnerability. To assess the broader relevance of the learned perturbation beyond this setting, we additionally evaluate cross-model and cross-dataset transfer in Section~\ref{sec:evaluation}. The attacker can optimize a universal patch directly against the victim model, fabricate the learned patch, and place it on the palm region during image acquisition. The same patch may be reused across multiple inputs and acquisition attempts.

\paragraph{Attack objective}
We consider both untargeted and targeted attacks. In the untargeted setting, the attacker aims to induce any incorrect identity prediction. In the targeted setting, the attacker aims to cause the recognizer to predict a specified target identity $y_t$.

\paragraph{Attack constraints and deployment setting}
The patch is constrained by a fixed pixel budget and a printable pixel range. The topology mask $M$ is fixed, whereas the texture $P$ is optimized. In our implementation, the universal patch is optimized on the training split of the dataset and then applied to unseen test images during evaluation. Most experiments are conducted in the digital domain under simulated acquisition effects, while additional physical-world experiments are performed to validate real-world feasibility. The attacker does not alter the victim model or the preprocessing pipeline other than applying the physical patch during image acquisition. Here, we focus on the robustness of the palmprint recognition model itself and do not explicitly consider liveness detection or multimodal authentication settings.




\section{Methodology}
\label{sec:method}

\begin{figure*}[t]
    \centering
    \includegraphics[width=\linewidth]{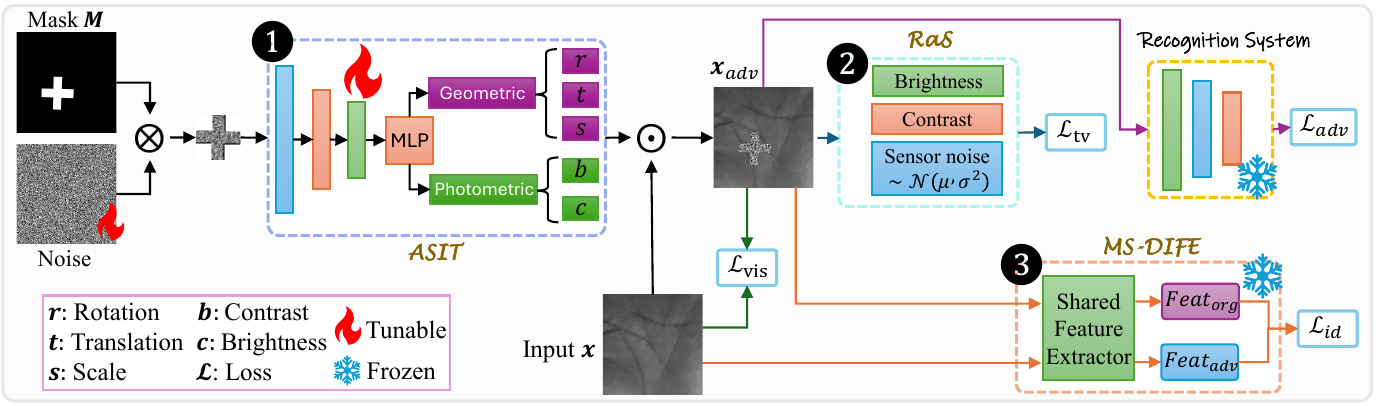}
    \caption{Training framework of \attacker. A universal cross-shaped patch, specified by a fixed mask $M$ and a learnable texture $P$, is rendered onto the input ROI through ASIT, which predicts input-conditioned rendering parameters. The composited sample is then processed by RaS to model capture-aware variation during training. In parallel, MS-DIFE extracts multi-scale features from the clean and rendered samples to provide the identity-related objective, while the victim recognizer provides the attack objective. The overall optimization is jointly driven by $\mathcal{L}_{\mathrm{adv}}$, $\mathcal{L}_{\mathrm{id}}$, $\mathcal{L}_{\mathrm{tv}}$, and $\mathcal{L}_{\mathrm{vis}}$.}    
    \label{fig:framework}
\end{figure*}

\subsection{Overview}

As illustrated in Fig.~\ref{fig:framework}, \attacker\ is a capture-aware universal adversarial patch framework against palmprint recognition systems. 
Let $x \in [0,1]^{H\times W}$ denote an aligned grayscale palmprint ROI image with identity label $y$. 
We learn a universal patch texture $P \in [0,1]^{H_p\times W_p}$ under a fixed cross-shaped binary mask $M \in \{0,1\}^{H_p\times W_p}$ and reuse the learned patch across different inputs and identities.

The central challenge is that a physically realizable perturbation must remain effective after print-and-capture variation, rather than only on digitally overlaid images. To address this challenge, \attacker\ adopts a differentiable rendering pipeline that combines input-conditioned patch adaptation with stochastic capture-aware simulation. 
Specifically, a fixed cross-shaped topology provides broad spatial coverage over discriminative palmprint structures, while the patch texture remains learnable.
ASIT then performs input-conditioned rendering adaptation to improve robustness to moderate variation in pose, scale, and local appearance. 
After rendering, RaS introduces stochastic capture-aware synthesis during training to improve robustness under practical acquisition conditions. In parallel, MS-DIFE provides frozen multi-scale feature guidance through an auxiliary branch, thereby strengthening identity-related feature disruption beyond the victim-space decision loss alone.

During inference, the adversarial sample is generated by a single forward rendering pass without further optimization and then directly evaluated by the victim recognizer.

\subsection{Cross-shaped Patch Topology}

Palmprint recognition relies heavily on ridge-and-line structures distributed over a broad spatial extent~\cite{PRL_22/EEPNet,PR_22/W2ML}. Under a fixed perturbation budget, a topology with broader spatial support is more likely to intersect principal palm lines and disturb long-range texture continuity than a compact block-like patch~\cite{TCSVT_24/CSPA}. Motivated by this observation, \attacker\ adopts a cross-shaped topology and fixes it through a binary mask $M$, while optimizing only the patch texture $P$.

Accordingly, the effective patch content is constrained by the masked texture
\begin{equation}
P_M = P \otimes M,
\label{eq:masked_patch}
\end{equation}
where $\otimes$ denotes element-wise multiplication. 

This formulation decouples \emph{topology} from \emph{appearance}: the cross-shaped support is fixed to preserve the desired spatial coverage, whereas the patch texture is learned to maximize attack effectiveness after the rendering and physical simulation.

\subsection{ASIT: Adaptive Spatial Transformer}

A physical patch attack is highly sensitive to input-dependent variation in pose, scale, and local appearance. If the perturbation is rendered in a rigid, input-agnostic manner, even mild acquisition variation may substantially reduce its effectiveness. We therefore introduce ASIT as an input-conditioned rendering module:
\begin{equation}
(\theta_{\mathrm{geo}},\theta_{\mathrm{pho}})
=
\mathrm{ASIT}_{\phi}(x),
\label{eq:asit}
\end{equation}
where $\phi$ denotes the learnable parameters of ASIT, $\theta_{\mathrm{geo}}$ specifies the geometric transformation parameters, and $\theta_{\mathrm{pho}}=(c,b)$ specifies a lightweight photometric calibration applied to the rendered patch before compositing. Here, $c$ and $b$ represent contrast-like scaling and brightness-like shifting factors, respectively.

The geometric component is parameterized by a low-dimensional affine transform,
\begin{equation}
\theta_{\mathrm{geo}}=(r,\mathbf{t},s),
\ 
\mathbf{t}=[t_x,t_y]^{\top},
\label{eq:asit_geo}
\end{equation}
where $r$ denotes the rotation angle, $\mathbf{t}$ denotes the 2D translation of the rendered patch within the ROI, and $s$ is an isotropic scale factor. The corresponding affine matrix is
\begin{equation}
A(\theta_{\mathrm{geo}})
=
\begin{bmatrix}
s\cos r & -s\sin r & t_x \\
s\sin r & \;\;s\cos r & t_y
\end{bmatrix}.
\label{eq:asit_affine}
\end{equation}
Rather than allowing unrestricted deformation, ASIT constrains the predicted transform to a bounded range around a physically plausible placement. This design improves robustness to moderate pose variation while avoiding unrealistic patch configurations that would be difficult to realize in practice.

Given the rendering parameters predicted for the current input, we warp both the masked patch and the binary support mask through a differentiable resampler:
\begin{equation}
\tilde{P}
=
\mathcal{W}_{\theta_{\mathrm{geo}}}(P \otimes M),
\ 
\tilde{M}
=
\mathcal{W}_{\theta_{\mathrm{geo}}}(M),
\label{eq:warp_patch}
\end{equation}
where $\mathcal{W}_{\theta_{\mathrm{geo}}}(\cdot)$ is implemented by differentiable grid sampling. The warped patch is then photometrically calibrated by
\begin{equation}
\bar{P}=c\tilde{P}+b,
\label{eq:pho_adjust}
\end{equation}
and composited onto the ROI as
\begin{equation}
\hat{x}
=
\mathcal{S}_{\theta_{\mathrm{geo}},\theta_{\mathrm{pho}}}(x,P,M)
=
(1-\tilde{M}) \otimes x + \tilde{M} \otimes \bar{P}.
\label{eq:compose}
\end{equation}
This rendering process is differentiable with respect to both the patch texture and the ASIT parameters, thereby enabling end-to-end optimization. Importantly, the photometric term in ASIT performs an input-conditioned local calibration of patch appearance before compositing, rather than stochastic scene-level augmentation.

\subsection{RaS: Radiometric Synthesis for Physical Robustness}
After differentiable compositing, we apply RaS to approximate the degradations introduced by print-and-capture acquisition. RaS operates on the entire composited ROI rather than only on the patch region, because a real sensor observes the full patched scene after the patch has been applied. Its role is therefore distinct from that of ASIT. Specifically, ASIT determines the primary input-conditioned rendering of the patch through geometric and patch-local photometric calibration, whereas RaS models residual stochastic variation at the scene level under practical acquisition conditions.

Taking the composited image $\hat{x}$ in \eqref{eq:compose} as input, we define the final rendered adversarial sample by
\begin{equation}
x_{\mathrm{adv}}(\xi)
=
\mathcal{A}_{\xi}(\hat{x}),
\ 
\xi \sim \mathcal{D}_{\mathrm{RaS}},
\label{eq:ras}
\end{equation}
where $\mathcal{D}_{\mathrm{RaS}}$ denotes the stochastic transformation distribution used to model practical acquisition variation. In practice, $\mathcal{A}_{\xi}(\cdot)$ captures perturbations such as photometric fluctuation and sensor noise. These transformations are deliberately kept lightweight, since their purpose is to simulate realistic acquisition uncertainty rather than dominate the rendering process.

This decomposition establishes a clear division of labor between the two modules. ASIT provides the principal input-conditioned refinement of patch placement and patch-local appearance before compositing, whereas RaS introduces stochastic scene-level variability that encourages robustness under physically plausible capture conditions. Following the expectation-over-transformations principle, the expectation over $\xi$ is approximated during training by Monte Carlo sampling, thereby exposing the optimization procedure to diverse realizations of the same underlying universal patch.

\subsection{MS-DIFE: Multi-Scale Feature Guidance}

Optimizing only the victim-space decision margin may be insufficient for palmprint attacks, since identity evidence is distributed across both fine-grained ridge textures and larger-scale line structures. To complement the victim-space objective, we introduce MS-DIFE as an auxiliary feature extractor that measures identity-related discrepancy across multiple spatial scales.

MS-DIFE adopts a Siamese-style formulation with shared weights for the clean input and the rendered adversarial sample. Let $E(\cdot)$ denote the shared encoder, and let $\hat{F}(\cdot)$ denote the corresponding recalibrated feature map after lightweight channel refinement. For the clean input $x$ and the rendered adversarial sample $x_{\mathrm{adv}}(\xi)$, we obtain $\hat{F}(x)$ and $\hat{F}(x_{\mathrm{adv}}(\xi))$, respectively. To capture identity-related structure at multiple resolutions, we aggregate each feature map by adaptive average pooling over a set of spatial scales $\mathcal{S}$. Specifically, we define
\begin{equation}
v(x)
=
\Big[
\mathrm{vec}\!\big(\Pi_s(\hat{F}(x))\big)
\Big]_{s\in\mathcal{S}},
\label{eq:msdife_desc_clean}
\end{equation}
and analogously
\begin{equation}
v\big(x_{\mathrm{adv}}(\xi)\big)
=
\Big[
\mathrm{vec}\!\big(\Pi_s(\hat{F}(x_{\mathrm{adv}}(\xi)))\big)
\Big]_{s\in\mathcal{S}},
\label{eq:msdife_desc_adv}
\end{equation}
where $\Pi_s(\cdot)$ denotes adaptive average pooling to an $s\times s$ grid, and $[\cdot]_{s\in\mathcal{S}}$ denotes concatenation over the selected scales. The final MS-DIFE embeddings are obtained by $\ell_2$ normalization:
\begin{equation}
g(x)=\frac{v(x)}{\|v(x)\|_2},
\qquad
g\big(x_{\mathrm{adv}}(\xi)\big)=\frac{v(x_{\mathrm{adv}}(\xi))}{\|v(x_{\mathrm{adv}}(\xi))\|_2}.
\label{eq:msdife_embed}
\end{equation}

MS-DIFE is pretrained on clean palmprint data and kept fixed during attack optimization. It provides a feature-space constraint that complements the victim-space attack loss by encouraging the adversarial sample to move away from the clean identity representation in the untargeted setting, or toward the target identity representation in the targeted setting. Such guidance is useful because the victim recognizer and the auxiliary feature extractor may emphasize different aspects of palmprint structure.

\subsection{Optimization}

We optimize \attacker\ under the EOT-based physical simulation pipeline by jointly minimizing an attack loss, an identity-related feature loss, a visual-consistency regularizer, and a total-variation regularizer.

\paragraph{Margin-based adversarial loss}
Let $z_j(\cdot)$ denote the victim score for class $j$. For the targeted setting with target identity $y_t$, we define
\begin{equation}
\small
\ell_{\mathrm{adv}}^{\mathrm{tar}}(x_{\mathrm{adv}},y_t)
=
\max\!\left\{
\max_{j\neq y_t} z_j(x_{\mathrm{adv}})
-
z_{y_t}(x_{\mathrm{adv}})
+
\kappa,\;
0
\right\},
\label{eq:cw_loss_targeted}
\end{equation}
where $\kappa \geq 0$ is the attack margin. For the untargeted setting with ground-truth identity $y$, we define
\begin{equation}
\small
\ell_{\mathrm{adv}}^{\mathrm{untar}}(x_{\mathrm{adv}},y)
=
\max\!\left\{
z_{y}(x_{\mathrm{adv}})
-
\max_{j\neq y} z_j(x_{\mathrm{adv}})
+
\kappa,\;
0
\right\}.
\label{eq:cw_loss_untargeted}
\end{equation}
Accordingly,
\begin{equation}
\mathcal{L}_{\mathrm{adv}}
=
\mathbb{E}_{(x,y)}
\mathbb{E}_{\xi\sim\mathcal{D}_{\mathrm{RaS}}}
\left[
\ell_{\mathrm{adv}}\big(x_{\mathrm{adv}}(\xi), y,y_t\big)
\right],
\end{equation}
where $\ell_{\mathrm{adv}}$ is instantiated as $\ell_{\mathrm{adv}}^{\mathrm{tar}}(\cdot,y_t)$ or $\ell_{\mathrm{adv}}^{\mathrm{untar}}(\cdot,y)$ according to the attack setting.

\paragraph{Identity-related feature loss}
To introduce feature-level guidance, we use the cosine distance
\begin{equation}
d_{\cos}(u,v)
=
1-\frac{\langle u,v\rangle}{\|u\|_2\|v\|_2}.
\label{eq:cos_dist}
\end{equation}
For untargeted attacks, we encourage the adversarial sample to move away from the clean identity representation:
\begin{equation}
\small
\mathcal{L}_{\mathrm{id}}^{\mathrm{untar}}
=
\mathbb{E}_{(x,y)}
\mathbb{E}_{\xi\sim\mathcal{D}_{\mathrm{RaS}}}
\left[
\max\!\left\{
0,\;
m-d_{\cos}\!\big(g(x),g(x_{\mathrm{adv}}(\xi))\big)
\right\}
\right],
\label{eq:id_loss_untar}
\end{equation}
where $m>0$ is an identity margin. For targeted attacks, we instead encourage the adversarial feature to approach a target prototype $g_t$:
\begin{equation}
\mathcal{L}_{\mathrm{id}}^{\mathrm{tar}}
=
\mathbb{E}_{(x,y)}
\mathbb{E}_{\xi\sim\mathcal{D}_{\mathrm{RaS}}}
\left[
d_{\cos}\!\big(g_t,g(x_{\mathrm{adv}}(\xi))\big)
\right].
\label{eq:id_loss_tar}
\end{equation}
We use $\mathcal{L}_{\mathrm{id}}$ to denote the corresponding identity term under the selected attack setting.

\paragraph{Total variation regularization}
To suppress high-frequency artifacts and improve printability, we regularize the patch texture by
\begin{equation}
\small
\mathcal{L}_{\mathrm{tv}}(P)
=
\sum_{u,v}
\left(
\|P_{u+1,v}-P_{u,v}\|_1
+
\|P_{u,v+1}-P_{u,v}\|_1
\right).
\label{eq:tv_loss}
\end{equation}

\paragraph{Visual-consistency regularization}
To prevent overly conspicuous rendering artifacts before stochastic synthesis, we regularize the composited image $\hat{x}$ against the clean ROI:
\begin{equation}
\mathcal{L}_{\mathrm{vis}}
=
\mathbb{E}_{(x,y)}
\left[
\|\hat{x}-x\|_2^2+\mathcal{L}_{\mathrm{ssim}}(\hat{x},x)
\right],
\label{eq:vis_loss}
\end{equation}
where $\mathcal{L}_{\mathrm{ssim}}(\cdot,\cdot)$ denotes the structural-similarity loss. This term acts before RaS and therefore constrains the rendered perturbation itself, rather than only its stochastically transformed realizations.

\paragraph{Overall objective}
The final optimization problem is

\begin{equation}
\min_{P,\phi}
\quad
\mathcal{L}
 =
\mathcal{L}_{\mathrm{adv}}
+
\lambda_{\mathrm{id}}\mathcal{L}_{\mathrm{id}}
+
\lambda_{\mathrm{tv}}\mathcal{L}_{\mathrm{tv}}(P)
+
\lambda_{\mathrm{vis}}\mathcal{L}_{\mathrm{vis}},
\label{eq:total_loss}
\end{equation}
The attack objective, feature-level identity guidance, and regularization terms are therefore optimized jointly under stochastic physical simulation. 
In practice, the expectation over $\xi$ is approximated by Monte Carlo sampling within each mini-batch during training, while MS-DIFE remains fixed throughout the optimization.

\begin{algorithm}[t]
\caption{\attacker\ Training}
\label{alg:training}
\begin{algorithmic}[1]
\Require Training set $\mathcal{D}_{\mathrm{train}}=\{(x_i,y_i)\}$, victim recognizer $f$, attack mode $s\in\{\text{targeted},\text{untargeted}\}$, target identity $y_t$ and target prototype $g_t$ if needed, number of training iterations $T$, mini-batch size $B$, number of EOT samples $K$
\Ensure Learned patch texture $P$, fixed topology mask $M$, and ASIT parameters $\phi$
\State Initialize universal patch texture $P$ and fixed topology mask $M$
\State Initialize ASIT parameters $\phi$
\For{$t=1$ to $T$}
    \State Sample a mini-batch $\mathcal{B}\subset \mathcal{D}_{\mathrm{train}}$ with $|\mathcal{B}|=B$
    \State Initialize $\mathcal{L}_{\mathrm{batch}} \gets 0$
    \ForAll{$(x,y)\in \mathcal{B}$}
        \State Predict rendering parameters $(\theta_{\mathrm{geo}},\theta_{\mathrm{pho}})=\mathrm{ASIT}_{\phi}(x)$
        \State Render the composited sample $\hat{x}=\mathcal{S}_{\theta_{\mathrm{geo}},\theta_{\mathrm{pho}}}(x,P,M)$
        \State Compute the per-sample visual-consistency loss on $(x,\hat{x})$
        \State Initialize $\mathcal{L}_{\mathrm{adv}}^{(x)} \gets 0$ and $\mathcal{L}_{\mathrm{id}}^{(x)} \gets 0$
        \For{$k=1$ to $K$}
            \State Sample $\xi_k \sim \mathcal{D}_{\mathrm{RaS}}$ and generate $x_{\mathrm{adv}}^{(k)}=\mathcal{A}_{\xi_k}(\hat{x})$
            \State Accumulate $\ell_{\mathrm{adv}}$ on $x_{\mathrm{adv}}^{(k)}$ 
            \State Accumulate the identity-related feature loss on $x_{\mathrm{adv}}^{(k)}$
        \EndFor
        \State Average over stochastic renderings:
        \[
        \mathcal{L}_{\mathrm{adv}}^{(x)} \gets \frac{1}{K}\mathcal{L}_{\mathrm{adv}}^{(x)},
        \ 
        \mathcal{L}_{\mathrm{id}}^{(x)} \gets \frac{1}{K}\mathcal{L}_{\mathrm{id}}^{(x)}
        \]
        \State Update the batch objective:
        \[
        \mathcal{L}_{\mathrm{batch}}
        \gets
        \mathcal{L}_{\mathrm{batch}}
        +
        \mathcal{L}_{\mathrm{adv}}^{(x)}
        +
        \lambda_{\mathrm{id}}\mathcal{L}_{\mathrm{id}}^{(x)}
        +
        \lambda_{\mathrm{vis}}\mathcal{L}_{\mathrm{vis}}(x,\hat{x})
        \]
    \EndFor
    \State Form the overall objective:
    \[
    \mathcal{L}_{\mathrm{batch}}
    \gets
    \frac{1}{B}\mathcal{L}_{\mathrm{batch}}
    +
    \lambda_{\mathrm{tv}}\mathcal{L}_{\mathrm{tv}}(P)
    \]
    \State Update $(P,\phi)$ by minimizing $\mathcal{L}_{\mathrm{batch}}$
    \State Project $P$ onto $[0,1]^{H_p\times W_p}$
\EndFor
\end{algorithmic}
\end{algorithm}

\subsection{Attacking}
\label{subsec:attacking}

\begin{figure}[t]
    \centering
    \includegraphics[width=0.98\linewidth]{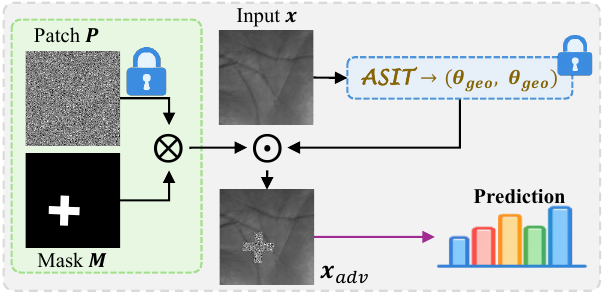}
    \caption{Attacking phase of \attacker. After training, the patch texture and the ASIT parameters are fixed. Given a test ROI image $x$, ASIT predicts the rendering parameters for the current input, and the adversarial sample is generated by a single forward rendering pass without test-time optimization. The resulting sample is then evaluated by the victim recognizer.}
    \label{fig:framework_test}
\end{figure}

After optimization, \attacker\ outputs the learned patch texture $P$, the fixed topology mask $M$, and the ASIT parameters $\phi$. During attacking, no further optimization is performed. As illustrated in Fig.~\ref{fig:framework_test}, given a new ROI image $x$, ASIT predicts the rendering parameters
\begin{equation}
(\theta_{\mathrm{geo}},\theta_{\mathrm{pho}})
=
\mathrm{ASIT}_{\phi}(x),
\label{eq:infer_asit}
\end{equation}
and the adversarial sample is generated by
\begin{equation}
x_{\mathrm{adv}}
=
\mathcal{S}_{\theta_{\mathrm{geo}},\theta_{\mathrm{pho}}}(x,P,M).
\label{eq:infer_comp}
\end{equation}
The resulting adversarial sample is then fed directly into the victim recognizer for evaluation. RaS is used during training to improve robustness to capture variation; at test time, the corresponding variability is provided either by the real acquisition process or by the evaluation protocol itself. This design keeps deployment simple: the learned perturbation remains universal, the test-time procedure is deterministic given the input ROI, and physical robustness is acquired during training rather than through additional online adaptation.

\begin{algorithm}[t]
\caption{\attacker\ Attacking}
\label{alg:attacking}
\begin{algorithmic}[1]
\Require Test set $\mathcal{D}_{\mathrm{test}}=\{(x_i,y_i)\}$, frozen patch texture $P$, fixed topology mask $M$, frozen ASIT parameters $\phi$, victim recognizer $f$
\Ensure Adversarial samples $\{x_i^{\mathrm{adv}}\}$ and corresponding victim predictions
\ForAll{$(x,y)\in \mathcal{D}_{\mathrm{test}}$}
    \State Predict rendering parameters $(\theta_{\mathrm{geo}},\theta_{\mathrm{pho}})=\mathrm{ASIT}_{\phi}(x)$
    \State Generate the adversarial sample
    \[
    x_{\mathrm{adv}}=\mathcal{S}_{\theta_{\mathrm{geo}},\theta_{\mathrm{pho}}}(x,P,M)
    \]
    \State Obtain the victim prediction $f(x_{\mathrm{adv}})$
\EndFor
\end{algorithmic}
\end{algorithm}


\section{Evaluation}
\label{sec:evaluation}
\subsection{Setup}
\label{subsec:setup}

\textbf{Datasets.}
We evaluate \attacker\ on two public palmprint datasets, Tongji~\cite{pr_17/tongji} and IITD~\cite{icvgip_08/iitd}, as well as AISEC, an in-house dataset collected from volunteer subjects. Informed consent was obtained from all participants prior to data collection, and the dataset will not be publicly released. Tongji and IITD serve as standardized benchmarks, whereas AISEC captures additional real-world variation. Tongji contains 300 subjects (600 palms) and 12,000 images, IITD contains 230 subjects (460 palms) and 2,300 ROI images, and AISEC contains 26 subjects (52 palms) and 1,040 images. During AISEC acquisition, each subject placed the hand flat on a desk, and images were captured from a top-down view using a smartphone under natural illumination at a distance of approximately 25--30 cm. For each palm, 20 images were collected and subsequently processed using ROI extraction, grayscale conversion, and Gaussian blurring. For each dataset, subjects are divided into disjoint training and test subsets. All samples are preprocessed into aligned $128\times128$ ROI images and, unless otherwise stated, all reported results are obtained on the test split.

\textbf{Models.}
We evaluate \attacker\ against a diverse set of victim models, including general-purpose CNN backbones (\textit{MobileNetV2}~\cite{cvpr_18/mobilenetv2}, \textit{VGG16}~\cite{ICLR_15/VGG}, \textit{ResNet-18}~\cite{cvpr_16/ResNet}, and \textit{ShuffleNetV2}~\cite{ECCV_18/shufflenet}) and palmprint-specific networks (\textit{CCNet}~\cite{tifs_23/ccnet}, \textit{CO3Net}~\cite{tim_23/CO3Net}, and \textit{CompNet}~\cite{spl_23/CompNet}). 
Across the evaluated datasets, these victim models attain near-saturated clean classification accuracy, indicating that the reported degradation is attributable to the attack rather than weak benign recognition.

\textbf{Baselines.}
We compare \attacker\ with representative patch-based attacks, including \textit{AdvPatch}~\cite{arXiv_17/AdvPatch}, two gradient-based patch variants implemented with \textit{MI-FGSM}~\cite{CVPR_18/MI-FGSM} and \textit{PGD}~\cite{iclr_18/PGD} (denoted as \textit{Patch$_{MI}$} and \textit{Patch$_{PGD}$}, respectively), as well as \textit{APPA}~\cite{TGRS_22/APPA}, \textit{AdvLogo}~\cite{arXiv_24/advlogo}, and \textit{CSPA}~\cite{TCSVT_24/CSPA}. In addition, we report a square-shaped variant of our method, denoted as \attacker$_s$, to isolate the effect of patch geometry. Unless otherwise specified, \attacker\ refers to the proposed cross-shaped version, denoted as \attacker$_c$.

\textbf{Implementation and evaluation.}
We jointly optimize the universal patch texture and the ASIT parameters using Adam with a learning rate of $5\times10^{-4}$. Unless otherwise specified, the regularization weights are set according to the sensitivity analysis in Section~\ref{subsec:ablation} as $\lambda_{\mathrm{id}}=0.20$, $\lambda_{\mathrm{vis}}=4\times10^{-3}$, and $\lambda_{\mathrm{tv}}=2\times10^{-5}$. For patch configuration, the square-patch baseline adopts a fixed size of $27\times27$ pixels. The proposed cross-shaped patch uses a long-arm length of $40$, while the short arm is fixed to $25\%$ of the long arm. Both patch variants are constrained to have comparable pixel budgets, ensuring a fair comparison. We report attack success rate (ASR, \%) as the evaluation metric. 
For untargeted attacks, ASR is computed over test samples that are correctly classified by the clean model and is defined as the fraction whose predictions change to any incorrect label after the attack. 
For targeted attacks, ASR is computed over test samples that are neither originally misclassified nor already assigned to the target identity by the clean model. It is defined as the fraction of such samples that are classified as the attacker-specified target identity after the attack.
All experiments are implemented in PyTorch and conducted on a Linux server equipped with $8\times$ NVIDIA H100 GPUs. 
In all tables, the best and second-best results are highlighted in \textbf{boldface} and \underline{underlining}, respectively, unless otherwise specified.

\subsection{Attack performance}
\subsubsection{Untargeted}


\begin{table}[t]
\centering
\small
\setlength{\tabcolsep}{2pt}
\renewcommand{\arraystretch}{1.15}
\caption{The untargeted attack success rate (\%) on IITD dataset.}
\resizebox{\linewidth}{!}{%
\begin{tabular}{lccccccc}
\toprule
Attack & VGG-16 & ResNet-18 & MobileNetV2 & ShuffleNetV2 & CompNet & CCNet & CO3Net \\
\midrule
AdvPatch    & 35.40 & 13.81 & 78.61 & 27.65 & 28.54 & 7.74  & 28.83 \\
Patch$_{MI}$     & 35.22 & 12.62 & 74.93 & 27.37 & 30.61 & 8.79  & 29.17 \\
Patch$_{PGD}$    & 35.58 & 12.75 & 70.96 & 30.87 & 30.03 & 8.79  & 29.28 \\
APPA          & 35.95 & 33.20 & 56.66 & 91.76 & 11.74 & 5.86  & 24.44 \\
CSPA          & 38.50 & 43.56 & 85.41 & \underline{98.04} & 25.89 & 9.61  & 25.56 \\
AdvLogo     & 97.30 & \textbf{95.28} & 73.66 & 97.94 & 66.39 & 2.93  & 3.60 \\
\attacker$_s$  & \textbf{98.18} & 70.65 & \underline{94.90} & 97.77 & \underline{67.09} & \underline{31.65} & \underline{63.63} \\
\attacker$_c$  & \underline{97.45} & \underline{88.71} & \textbf{96.46} & \textbf{98.74} & \textbf{92.98} & \textbf{79.48} & \textbf{87.39} \\
\bottomrule
\end{tabular}
}
\label{tab:untarget_iitd}
\end{table}


\begin{table}[t]
\centering
\small
\setlength{\tabcolsep}{2pt}
\renewcommand{\arraystretch}{1.15}
\caption{The untargeted attack success rate (\%) on Tongji dataset.}
\resizebox{\linewidth}{!}{%
\begin{tabular}{lccccccc}
\toprule
Attack & VGG-16 & ResNet-18 & MobileNetV2 & ShuffleNetV2 & CompNet & CCNet & CO3Net \\
\midrule
AdvPatch    & 80.39 & \textbf{100.00} & 82.50 & 90.91 & 49.36 & 85.77 & 91.32 \\
Patch$_{MI}$     & 80.39 & \textbf{100.00} & 82.50 & 45.45 & 52.09 & 81.63 & 72.34 \\
Patch$_{PGD}$    & 90.20 & 28.57 & 82.50 & 45.45 & 51.36 & 80.84 & 87.58 \\
APPA          & 82.35 & 68.57 & 75.00 & 81.82 & 47.51 & 60.72 & 73.53 \\
CSPA          & 98.75 & \underline{98.92} & 96.03 & \textbf{95.93} & 59.00 & 94.23 & 89.45 \\
AdvLogo     & 54.69 & 51.04 & 87.70 & 53.42 & 22.92 & 18.41 & 35.83 \\
\attacker$_s$  & \underline{99.28} & 95.69 & \underline{98.77} & 88.88 & \underline{97.68} & \underline{99.55} & \underline{98.46} \\
\attacker$_c$  & \textbf{99.60} & 95.95 & \textbf{99.03} & \underline{95.52} & \textbf{99.80} & \textbf{99.83} & \textbf{99.60} \\
\bottomrule
\end{tabular}
}
\label{tab:untarget_tongji}
\end{table}


\begin{table}[t]
\centering
\small
\setlength{\tabcolsep}{2pt}
\renewcommand{\arraystretch}{1.15}
\caption{The untargeted attack success rate (\%) on AISEC dataset.}
\resizebox{\linewidth}{!}{%
\begin{tabular}{lccccccc}
\toprule
Attack & VGG-16 & ResNet-18 & MobileNetV2 & ShuffleNetV2 & CompNet & CCNet & CO3Net \\
\midrule
AdvPatch    & 31.42 & 87.84 & 48.32 & \textbf{97.90} & 67.16 & 24.42 & 92.37 \\
Patch$_{MI}$     & 29.51 & 89.52 & 40.13 & \textbf{97.90} & 66.32 & 20.42 & 92.80 \\
Patch$_{PGD}$    & 21.02 & 20.63 & 69.54 & 25.46 & 65.47 & 19.79 & 92.37 \\
APPA          & 95.33 & 30.61 & 10.08 & \underline{97.06} & 53.89 & 14.00 & 87.97 \\
CSPA          & 94.06 & \textbf{97.90} & 22.48 & \underline{97.74} & 87.79 & 23.37 & \underline{95.97} \\
AdvLogo     & 97.88 & \underline{97.48} & \textbf{73.66} & \textbf{97.90} & 12.42 & 10.95 & 72.88 \\
\attacker$_s$  & \underline{99.36} & 96.02 & 48.32 & 92.23 & \underline{97.68} & \underline{42.68} & 93.43 \\
\attacker$_c$  & \textbf{99.79} & \textbf{97.90} & \underline{71.64} & 95.59 & \textbf{99.16} & \textbf{88.42} & \textbf{97.88} \\
\bottomrule
\end{tabular}
}
\label{tab:untarget_aisec}
\end{table}

We first evaluate \emph{untargeted} attacks, where the adversary aims to induce any incorrect identity prediction. 
Tables~\ref{tab:untarget_iitd}--\ref{tab:untarget_aisec} show that the \attacker\ family achieves the strongest overall untargeted performance across datasets and victim architectures, with \attacker$_c$ providing the most reliable results. 
Its advantage lies not only in higher mean ASR but also in stronger consistency across heterogeneous victims. In particular, \attacker$_c$ attains the highest average ASR over the seven evaluated models on all three datasets, namely 92.91\% on AISEC, 91.60\% on IITD, and 98.48\% on Tongji.

This advantage is most visible on AISEC and IITD, where the comparison is more diagnostic. Several competing attacks perform well on a subset of generic CNN backbones, yet deteriorate sharply on palmprint-specific models. By contrast, \attacker$_c$ remains strong on both model families, indicating that the learned perturbation is less tied to the inductive bias of a particular recognizer. 
This distinction is practically important because the deployed victim architecture is often unknown.

A closer look at the per-model results supports this interpretation. 
On AISEC, Patch$_{MI}$, Patch$_{PGD}$, and \textit{AdvLogo} all exhibit pronounced instability on at least one palmprint-specific target, whereas \attacker$_c$ maintains high ASR simultaneously on CompNet, CCNet, and CO3Net. On IITD, \textit{AdvLogo} is near-saturated on several generic CNNs but drops to 2.93\% and 3.60\% on CCNet and CO3Net, respectively, while \attacker$_c$ remains at 79.48\% and 87.39\%. These gaps indicate that many existing baselines still rely heavily on architecture-specific attack cues, whereas \attacker$_c$ transfers more effectively across model families.

Tongji appears less challenging under the present protocol, as many methods achieve higher ASR. However, this does not eliminate the separation between methods. Even in this higher-ASR regime, \attacker$_c$ is the only method that remains above 95\% on all seven architectures, which indicates that its advantage is not merely a consequence of favorable dataset conditions, but of stronger cross-architecture stability.

Overall, the untargeted results show that \attacker, especially \attacker$_c$, combines high average ASR with strong worst-case performance across victim models. This makes it a more reliable attacker under heterogeneous-victim uncertainty and therefore a stronger tool for practical threat assessment.

\subsubsection{Targeted}


\begin{table}[ht]
\centering
\small
\setlength{\tabcolsep}{2pt}
\renewcommand{\arraystretch}{1.15}
\caption{The targeted attack success rate (\%) on IITD dataset.}
\resizebox{\linewidth}{!}{%
\begin{tabular}{lccccccc}
\toprule
Attack & VGG-16 & ResNet-18 & MobileNetV2 & ShuffleNetV2 & CompNet & CCNet & CO3Net \\
\midrule
AdvPatch & 0.91 & 0.53 & 12.48 & 2.10 & 4.73 & 3.17 & 20.20 \\
Patch$_{MI}$    & 0.91 & 0.40 & 6.24  & 0.84 & 0.81 & 4.23 & 20.88 \\
Patch$_{PGD}$        & 0.91 & 0.53 & 11.49 & 1.54 & 1.04 & 5.16 & 20.65 \\
APPA       & 0.91 & 0.13 & 5.82  & 7.42 & 0.69 & 0.70 & 0.23 \\
AdvLogo  & 0.91 & 3.73 & 12.77 & \underline{21.57} & 67.94 & 13.50 & 31.60 \\
CSPA       & 1.46 & 4.66 & 23.69 & \textbf{27.03} & 80.62 & 23.59 & 76.64 \\
\attacker$_s$   & \textbf{3.66} & \underline{15.45} & \textbf{30.21} & 15.83 & \underline{99.31} & \underline{72.89} & \underline{98.42} \\
\attacker$_c$   & \underline{1.83} & \textbf{17.31} & \underline{25.11} & 16.53 & \textbf{99.65} & \textbf{86.38} & \textbf{99.44} \\
\bottomrule
\end{tabular}
}
\label{tab:targeted_iitd}
\end{table}


\begin{table}[ht]
\centering
\small
\setlength{\tabcolsep}{2pt}
\renewcommand{\arraystretch}{1.15}
\caption{The targeted attack success rate (\%) on Tongji dataset. }
\resizebox{\linewidth}{!}{%
\begin{tabular}{lccccccc}
\toprule
Attack & VGG-16 & ResNet-18 & MobileNetV2 & ShuffleNetV2 & CompNet & CCNet & CO3Net \\
\midrule
AdvPatch & 0.29 & 11.53 & 8.59  & 17.67 & 78.31 & 49.29 & 99.51 \\
Patch$_{MI}$    & 0.05 & 0.94  & 2.91  & 11.39 & 74.60 & 67.61 & 99.33 \\
Patch$_{PGD}$        & 0.27 & 11.50 & 8.71  & 17.06 & 77.89 & 71.16 & \underline{99.75} \\
APPA       & 0.00 & 0.00  & 1.54  & 4.81  & 32.62 & 52.91 & 98.83 \\
AdvLogo  & 7.97 & 10.35 & 6.77  & 20.52 & 95.58 & 97.96 & \textbf{100.00} \\

CSPA       & \underline{9.70} & 18.78 & 13.78 & \underline{36.05} & \underline{96.37} & \underline{99.68} & \textbf{100.00} \\
\attacker$_s$   & 1.54 & \underline{41.04} & \underline{42.85} & 33.84 & \textbf{100.00} & \textbf{100.00} & \textbf{100.00} \\
\attacker$_c$   & \textbf{10.44} & \textbf{46.52} & \textbf{61.15} & \textbf{74.77} & \textbf{100.00} & \textbf{100.00} & \textbf{100.00} \\
\bottomrule
\end{tabular}
}
\label{tab:targeted_tongji}
\end{table}


\begin{table}[ht]
\centering
\small
\setlength{\tabcolsep}{2pt}
\renewcommand{\arraystretch}{1.15}
\caption{The targeted attack success rate (\%) on AISEC dataset.}
\resizebox{\linewidth}{!}{%
\begin{tabular}{lccccccc}
\toprule
Attack & VGG-16 & ResNet-18 & MobileNetV2 & ShuffleNetV2 & CompNet & CCNet & CO3Net \\
\midrule
AdvPatch & 0.00 & 0.21 & 0.63 & 0.63 & 34.95 & 0.00 & 79.87 \\
Patch$_{MI}$    & \textbf{1.27} & 0.42 & 1.89 & 2.52 & 38.53 & 0.00 & 91.95 \\
Patch$_{PGD}$        & 0.00 & 0.21 & 0.84 & 0.42 & 37.68 & 0.00 & 90.89 \\
APPA       & \textbf{1.27} & 0.00 & \underline{2.10} & 0.84 & 7.16 & 0.21 & 66.31 \\
AdvLogo  & 0.42 & \underline{1.47} & \textbf{3.36} & 7.77 & 78.74 & 1.89 & 94.49 \\
CSPA       & \underline{1.06} & \textbf{2.31} & \textbf{3.36} & \textbf{14.50} & \underline{90.11} & 38.53 & 98.52 \\
\attacker$_s$   & \textbf{1.27} & 0.00 & \textbf{3.36} & 7.98 & \textbf{100.00} & \underline{78.32} & \underline{99.79} \\
\attacker$_c$   & 0.64 & 0.00 & 0.84 & \underline{8.82} & \textbf{100.00} & \textbf{83.79} & \textbf{100.00} \\
\bottomrule
\end{tabular}
}
\label{tab:targeted_aisec}
\end{table}

We further evaluate \emph{targeted} attacks, where the adversary aims to force the victim to predict a pre-specified target identity. Throughout this section, the target label is fixed to $0$. Compared with untargeted attacks, targeted attacks are more demanding because they require not only suppressing the true identity but also steering the prediction toward a specific incorrect identity. As a result, targeted ASR is generally more sensitive to model architecture and dataset characteristics.

Across IITD and AISEC (Tables~\ref{tab:targeted_iitd} and \ref{tab:targeted_aisec}), most prior baselines exhibit a clear model-family gap: they may achieve nontrivial targeted ASR on some generic CNN backbones, yet fail to reliably control palmprint-specific models. Gradient-based patch variants remain particularly weak under the targeted objective, indicating that directly optimizing a generic patch loss is insufficient to consistently steer predictions toward a fixed target identity. Representative patch-based baselines improve targeted success on some architectures, but still show pronounced brittleness across victims.

Our method substantially reduces this brittleness. On both IITD and AISEC, \attacker\ maintains strong targeted performance on palmprint-specific models and provides a markedly more stable operating regime than competing attacks. In particular, the cross-shaped variant \attacker$_c$ is the most reliable method on the palmprint-specific victims. For example, on IITD it achieves 99.65\%, 86.38\%, and 99.44\% ASR on CompNet, CCNet, and CO3Net, respectively, and on AISEC it reaches 100.00\%, 83.79\%, and 100.00\% on the same three models. This pattern suggests that the cross-shaped geometry is more effective for targeted manipulation in palmprint recognition, especially on palmprint-specific models.
This pattern suggests that the cross-shaped geometry is better aligned with the targeted objective, since it more effectively perturbs the texture continuity cues that dominate palmprint recognition while still imprinting target-oriented patterns.

On Tongji (Table~\ref{tab:targeted_tongji}), targeted ASR is uniformly higher for most methods, and several approaches are close to saturation on palmprint-specific models. We therefore interpret Tongji mainly as evidence that targeted steering is feasible on a comparatively easier dataset, while the more diagnostic separation remains on the generic CNN backbones. Under this view, \attacker$_c$ still stands out as the strongest and most consistent option across all four CNN models, indicating that its advantage is not simply due to easier data, but to stronger controllability across heterogeneous victims.

Overall, the targeted experiments support two conclusions. First, existing attacks remain strongly architecture-dependent under targeted objectives, especially on palmprint-specific recognizers. Second, \attacker, particularly \attacker$_c$, provides superior targeted controllability together with stronger cross-architecture consistency, making it a more informative tool for evaluating worst-case targeted vulnerability in practical palmprint recognition systems.

\subsection{Transferability across models}


\paragraph{Pairwise transfer}
\begin{figure}[ht]
    \centering
    \includegraphics[width=\linewidth]{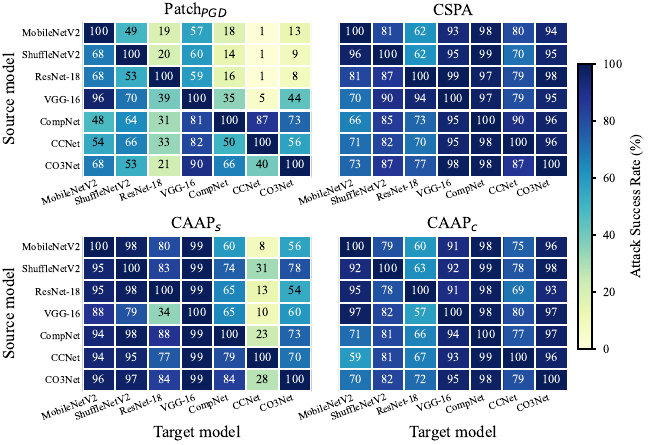}
    \caption{Pairwise cross-model transferability. Each heatmap reports ASR (\%) when a patch crafted on a source model (rows) is directly transferred to a target model (columns).}
    \label{fig:transfer_heatmap}
\end{figure}

We evaluate cross-model transferability by generating an adversarial patch on a source model and directly applying it to a different target model. Fig.~\ref{fig:transfer_heatmap} summarizes the resulting ASR across seven architectures under four representative methods, namely Patch$_{PGD}$, \textit{CSPA}, \attacker$_s$, and \attacker$_c$. 
The key observation is that \attacker$_c$ exhibits the strongest and most consistent \emph{off-diagonal} transfer pattern, indicating that the learned cross-shaped patch is less prone to overfitting to the source model.
\attacker$_s$ also transfers well to several targets, but it degrades more noticeably on some palmprint-specific victims, especially CCNet, suggesting that patch geometry affects not only white-box strength but also transfer stability. In contrast, Patch$_{PGD}$ shows the weakest and most uneven transferability, which is consistent with stronger dependence on source-specific gradients. \textit{CSPA} improves over Patch$_{PGD}$ on many source-target pairs, yet still exhibits non-negligible gaps on more difficult combinations. Overall, these results indicate that the proposed \attacker\ design, particularly the cross-shaped variant, improves cross-model generalization and is therefore better suited to black-box settings in which the deployed target model differs from the source used during patch crafting.


\paragraph{Hold-out transfer}
We next consider a more stringent \emph{hold-out} transfer setting, where the victim model is excluded from patch crafting. Specifically, adversarial patches are crafted using an ensemble of six source models and then directly evaluated on a single unseen target model. This setting imposes a stronger test of architecture-level generalization, since no target-specific information is available during optimization. Table~\ref{tab:holdout} reports the resulting ASR across seven hold-out targets. The results show the same overall trend: both \attacker\ variants transfer substantially better than the gradient-based and prior patch-based baselines, and \attacker$_c$ is the strongest method on six of the seven targets. The gains are especially pronounced on CompNet and CCNet, where \attacker$_c$ reaches 98.71\% and 79.25\%, compared with 12.22\% and 0.17\% for Patch$_{PGD}$. \attacker$_s$ is also competitive and performs best on VGG-16, indicating that patch geometry can influence transfer differently across architectures. 
Overall, the hold-out setting leads to the same conclusion as the pairwise study: \attacker\ transfers more effectively across architectures and is therefore better suited to black-box deployment where the victim model is unavailable during optimization.

\begin{table}[t]
\centering
\small
\setlength{\tabcolsep}{2pt}
\caption{Hold-out transfer ASR (\%). Adversarial patches are crafted using an ensemble of six source models and directly evaluated on a single unseen target model.}
\label{tab:holdout}
\resizebox{\linewidth}{!}{%
\begin{tabular}{lccccccc}
\toprule
Method & MobileNetV2 & ShuffleNetV2 & ResNet-18 & VGG-16 & CompNet & CCNet & CO3Net \\
\midrule
Patch$_{PGD}$           & 22.17             & 30.17             & 73.52             & 87.05             & 12.22             & 0.17              & 13.61             \\
CSPA          & 43.75             & 83.67             & \underline{84.75} & \underline{96.28} & \underline{96.98} & \underline{41.45} & \underline{96.60} \\
\attacker$_s$ & \underline{91.03} & \underline{96.15} & 83.47             & \textbf{99.18}    & 60.98             & 7.95              & 72.49             \\
\attacker$_c$  & \textbf{92.97}    & \textbf{96.53}    & \textbf{85.30}    & 89.50             & \textbf{98.71}    & \textbf{79.25}    & \textbf{96.70}    \\
\bottomrule
\end{tabular}
}
\end{table}

\subsection{Transferability across datasets}
We further evaluate \emph{cross-dataset} generalization by training \attacker\ on Tongji and then directly applying the learned universal patch and ASIT module to IITD, without any additional fine-tuning or calibration on IITD.
As reported in Table~\ref{tab:tongji_to_iitd_transfer}, both \attacker\ variants remain effective under this dataset shift, achieving consistently high ASR across diverse target architectures. The pattern is also informative at the model level: \attacker$_s$ performs best on MobileNetV2 and ShuffleNetV2, whereas \attacker$_c$ performs best on ResNet-18, VGG-16, CompNet, and CO3Net, while remaining competitive on CCNet. In contrast, Patch$_{PGD}$ and \textit{CSPA} exhibit substantially lower ASR on most targets under the same protocol. 
These results suggest that \attacker\ is relatively robust to changes in subject identities and acquisition conditions under cross-dataset transfer.

\begin{table}[t]
\centering
\small
\setlength{\tabcolsep}{3pt}
\caption{Cross-dataset transfer ASR (\%) from Tongji to IITD.}
\label{tab:tongji_to_iitd_transfer}
\resizebox{\linewidth}{!}{%
\begin{tabular}{lccccccc}
\toprule
Method & MobileNetV2 & ShuffleNetV2 & ResNet-18 & VGG-16 & CompNet & CCNet & CO3Net \\
\midrule
Patch$_{PGD}$           & 62.41 & 61.76 & 31.56 & 50.27 & 17.19 & 7.17  & 32.96 \\
CSPA          & 73.48 & 54.62 & 24.10 & 62.34 & 20.88 & 7.76  & 22.69 \\
\attacker$_s$ & \textbf{97.73} & \textbf{96.79} & \underline{86.59} & \underline{88.50} & \underline{60.07} & \textbf{28.02} & \underline{74.21} \\
\attacker$_c$  & \underline{93.77} & \underline{86.31} & \textbf{92.83} & \textbf{99.27} & \textbf{69.51} & \underline{27.78} & \textbf{87.16} \\
\bottomrule
\end{tabular}
}
\end{table}


\subsection{Adversarial Training}

\begin{figure}[t]
    \centering
    \includegraphics[width=0.9\linewidth]{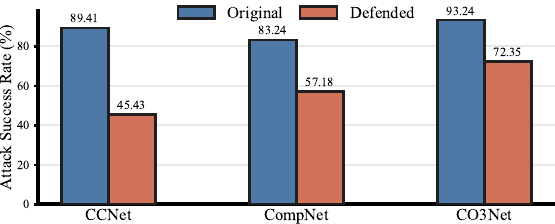}
    \caption{ASR of \attacker\ on three palmprint-specific recognizers before and after adversarial training.}
    \label{fig:defense}
\end{figure}

We further evaluate whether adversarial training mitigates \attacker\ under a practical deployment setting. Specifically, we adversarially train three palmprint-specific recognizers using optimized \attacker\ patches, and then re-optimize the attack patch against the defended models and re-evaluate ASR under the same protocol. Fig.~\ref{fig:defense} reports the ASR before and after defense.

Overall, adversarial training consistently reduces the effectiveness of \attacker\ across all three models, indicating that training-time hardening can suppress a substantial portion of the attack signal. However, the reduction is only partial: the defended ASR remains non-negligible on all three architectures, and the magnitude of the reduction is clearly model-dependent. This suggests that the robustness gained from adversarial training depends on how each recognizer encodes local texture and geometric cues, and that a single defense recipe may not provide uniform protection across different palmprint recognition pipelines.

These observations indicate that adversarial training should be interpreted as a partial mitigation rather than a complete solution against \attacker-style patch attacks. The residual attack success therefore motivates the development of more palmprint-specific defense strategies against structured physical perturbations.

\subsection{Physical Attack}

\begin{figure}[t]
    \centering
    \includegraphics[width=0.9\linewidth]{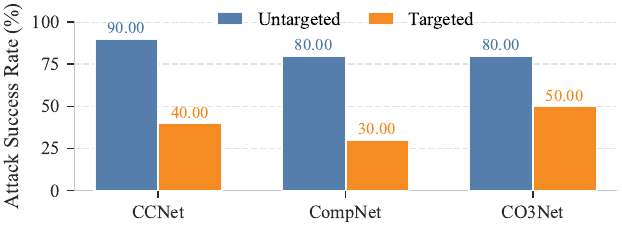}
    \caption{Identity-level ASR of physical untargeted and targeted attacks by \attacker\ under real print-and-capture acquisition.}    
    \label{fig:physical}
\end{figure}

To validate the practicality of \attacker\ beyond simulation, we conduct physical attack experiments on AISEC under both the untargeted and targeted settings. The optimized patch is scaled to the target physical size and printed in two forms: one binary black-and-white version and five randomly sampled RGB realizations with the same grayscale appearance. For each subject and each attack setting, we collect 20 physical attack images, including 10 captured with the black-and-white patch and 10 captured with the sampled RGB realizations, under capture conditions matched as closely as possible to AISEC data collection. Using 10 subjects, this yields 200 physical attack images for the untargeted setting and 200 for the targeted setting, resulting in 400 physical attack images overall. 
To ensure consistency with the AISEC pipeline, we apply the same preprocessing procedure used in AISEC data construction before feeding it to the victim models (as described in Sec.~\ref{subsec:setup}).

We report identity-level physical attack success separately for the untargeted and targeted settings. In the untargeted, an identity is regarded as successfully attacked if at least one of its 20 physical attack images induces misclassification. In the targeted, success requires that at least one of the 20 physical attack images be classified as the designated target identity. 

As shown in Fig.~\ref{fig:physical}, \attacker\ maintains high untargeted physical attack success across victim models, indicating that the attack learned under the capture-aware simulation pipeline transfers effectively to real print-and-capture conditions. Targeted physical attacks are more difficult, yet they still achieve nontrivial success, showing that current palmprint recognizers remain vulnerable even when the perturbation must survive printing, attachment, re-capture, and ROI preprocessing. These results therefore support the physical transferability of the proposed attack under real acquisition conditions.

\subsection{Ablation Study}
\label{subsec:ablation}
\subsubsection{Size}
\begin{figure}[t]
    \centering
    \includegraphics[width=0.8\linewidth]{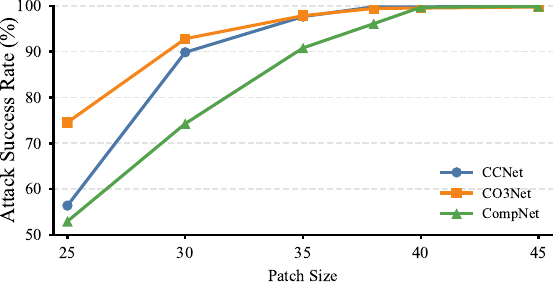}
    \caption{Effect of cross-shaped patch size. ASR (\%) is reported against CCNet, CO3Net, and CompNet as the long-arm length varies from 25 to 45, with the short arm fixed to 25\% of the long arm.}
    \label{fig:patch_size}
\end{figure}
We study the impact of patch size for the cross-shaped design by sweeping the long-arm length from 25 to 45 while fixing the short arm to 25\% of the long arm. As shown in Fig.~\ref{fig:patch_size}, increasing the patch size consistently improves ASR across all three palmprint-specific models, although the rate of improvement differs by architecture. CCNet and CO3Net improve rapidly from 25 to 30 and then approach saturation, whereas CompNet improves more gradually and requires a larger size to reach a comparable regime. This pattern suggests that a larger cross-shaped support is more effective for disrupting palmprint recognizers. Based on this trade-off, we adopt a long-arm length of 40 as the default configuration, since it delivers stable near-saturated performance without requiring a larger patch.

\subsubsection{Shape}
\begin{figure}[t]
    \centering
    \includegraphics[width=\linewidth]{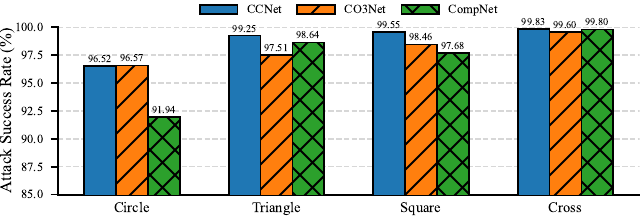}
    \caption{Ablation on patch shape. ASR (\%) is reported for four patch geometries against three palmprint-specific recognizers.}
    \label{fig:patch_shape}
\end{figure}

We further study the impact of patch shape by comparing four representative designs, namely square, circle, triangle, and cross, against three palmprint-specific models. As shown in Fig.~\ref{fig:patch_shape}, the cross-shaped patch achieves consistently high ASR across all models, indicating that a sparse, structure-aware geometry is more effective under the present attack setting. The triangle patch is also competitive, whereas the square patch shows a noticeable drop on CompNet, suggesting that a compact and uniform geometry is less aligned with the critical feature responses of that matcher under our attack setup. Overall, these results indicate that patch geometry materially affects attack effectiveness. We therefore use the cross-shaped design in subsequent experiments.

\subsubsection{Position}

\begin{figure}[t]
    \centering
    \includegraphics[width=\linewidth]{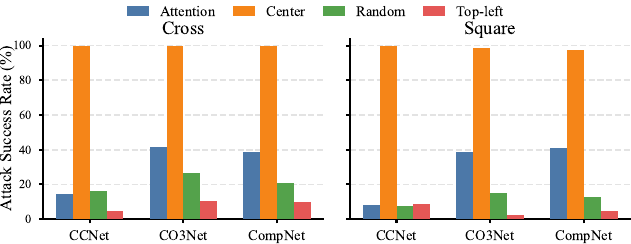}
    \caption{Effect of patch placement position. ASR (\%) is reported for attention-guided, center, random, and fixed top-left placement.}
    
    \label{fig:patch_position}
\end{figure}

We further conduct an ablation study on patch placement by evaluating four representative positions: attention-guided placement, center placement, random placement, and fixed top-left placement. As shown in Fig.~\ref{fig:patch_position}, center placement consistently yields the highest ASR across all three palmprint-specific models for both cross- and square-shaped patches, suggesting that the central ROI region is a particularly effective placement location under the present setting. In contrast, random and corner placements lead to markedly lower success rates, suggesting that misaligned patch locations often fail to interfere with the most discriminative regions. Attention-guided placement improves over random and corner placement, but still trails center placement, suggesting that the current attention proxy is less reliable than the simple central prior for identifying the most effective attack region. We therefore adopt center placement as the default strategy.

\subsubsection{Components}
Table~\ref{tab:ablation_components} shows that the three proposed components are complementary and that the full design yields the strongest overall performance. 
The base variant, which removes ASIT, MS-DIFE, and RaS, achieves only 63.79\%, 57.59\%, and 35.79\% ASR on CCNet, CO3Net, and CompNet, respectively. Enabling ASIT alone yields the largest improvement, indicating that geometry-aware patch rendering is the primary driver of attack strength in our framework. By contrast, MS-DIFE or RaS alone offers only limited benefit over the base variant, which suggests that feature-level guidance or radiometric augmentation is insufficient without strong rendering adaptation. 
Once combined with ASIT, however, these components provide further gains, and the full model achieves the best results on all victim models. 
Overall, the ablation indicates that ASIT provides the primary gain, whereas MS-DIFE and RaS act as complementary refinements that further improve performance within the complete framework.

\begin{table}[t]
\centering
\small
\setlength{\tabcolsep}{5pt}
\caption{Ablation on component combinations. Base removes ASIT, MS-DIFE, and RaS, whereas All enables all three components.}
\label{tab:ablation_components}
\resizebox{\linewidth}{!}{%
\begin{tabular}{lccc|ccc}
\toprule
\multirow{2}{*}{Setting} & \multicolumn{3}{c|}{Components} & \multicolumn{3}{c}{Victim model} \\
\cmidrule(lr){2-4}\cmidrule(lr){5-7}
 & ASIT & MS-DIFE & RaS & CCNet & CO3Net & CompNet \\
\midrule
Base          &            &            &            & 63.79 & 57.59 & 35.79 \\
\hdashline
ASIT          & \checkmark &            &            & 97.43 & 97.52 & 89.52 \\
MS-DIFE       &            & \checkmark &            & 62.24 & 57.54 & 36.21 \\
RaS           &            &            & \checkmark & 65.46 & 57.29 & 35.94 \\
ASIT+MS-DIFE  & \checkmark & \checkmark &            & 97.82 & 97.47 & 84.09 \\
ASIT+RaS      & \checkmark &            & \checkmark & 97.75 & 97.37 & 84.53 \\
MS-DIFE+RaS   &            & \checkmark & \checkmark & 61.79 & 56.52 & 35.69 \\
Full          & \checkmark & \checkmark & \checkmark & \textbf{99.83} & \textbf{99.60} & \textbf{99.80} \\
\bottomrule
\end{tabular}%
}

\end{table}

\subsubsection{Hyper-parameter sensitivity}
\label{subsubsec:parameter}
\begin{figure*}[ht]
    \centering
    \includegraphics[width=\linewidth]{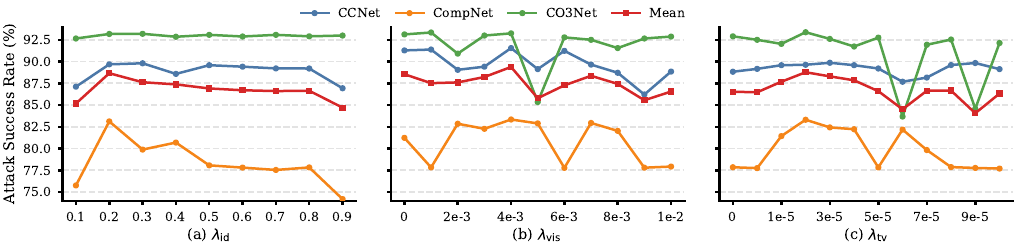}  
    \caption{Hyper-parameter sensitivity on Tongji. Untargeted ASR (\%) is reported for CCNet, CompNet, and CO3Net, together with their mean, under one-at-a-time sweeps of $\lambda_{\mathrm{id}}$, $\lambda_{\mathrm{vis}}$, and $\lambda_{\mathrm{tv}}$. \textit{Note:} The patch size is set to 30 in this ablation study to make the differences across settings more discernible.}
    \label{fig:hypers_2}
\end{figure*}
We further examine the sensitivity of the objective on Tongji by sweeping $\lambda_{\mathrm{id}}$, $\lambda_{\mathrm{vis}}$, and $\lambda_{\mathrm{tv}}$, while evaluating three representative palmprint models, namely CCNet, CompNet, and CO3Net, together with their mean ASR, as shown in Fig.~\ref{fig:hypers_2}. In each sweep, only one hyper-parameter is varied while the other two are fixed. 

\paragraph{Effect of $\lambda_{\mathrm{id}}$}
Fig.~\ref{fig:hypers_2}(a) shows that $\lambda_{\mathrm{id}}$ exhibits a relatively broad high-performing region. 
The mean ASR reaches its maximum at $\lambda_{\mathrm{id}}=0.2$ and remains comparatively stable over a wide intermediate range. The degradation at overly large values is mainly driven by CompNet and CCNet, whereas CO3Net remains near-saturated throughout the sweep. These results suggest that an excessively large identity-related weight can over-constrain the optimization on some palmprint-specific backbones, while a moderate value provides a better balance between average performance and cross-model stability.

\paragraph{Effect of $\lambda_{\mathrm{vis}}$}
Fig.~\ref{fig:hypers_2}(b) indicates that the visual-consistency term also requires moderate weighting. When $\lambda_{\mathrm{vis}}$ is too small, the mean ASR remains competitive but does not reach its best level, suggesting that insufficient appearance regularization may leave visible rendering artifacts insufficiently controlled.
As $\lambda_{\mathrm{vis}}$ increases, the mean ASR improves and reaches its best value at $\lambda_{\mathrm{vis}}=4\times10^{-3}$. Beyond this point, the performance becomes non-monotonic and shows noticeable drops, indicating that overly strong visual regularization can undesirably restrict the optimization. Overall, the sweep supports choosing $\lambda_{\mathrm{vis}}=4\times10^{-3}$ as a robust operating point.

\paragraph{Effect of $\lambda_{\mathrm{tv}}$}
Fig.~\ref{fig:hypers_2}(c) shows that light total-variation regularization is beneficial, whereas overly large $\lambda_{\mathrm{tv}}$ can cause sharp and non-monotonic degradation. The mean ASR reaches its maximum at $\lambda_{\mathrm{tv}}=2\times10^{-5}$, which suggests that mild smoothness constraints suppress spurious artifacts without overly restricting the adversarial objective. When $\lambda_{\mathrm{tv}}$ becomes larger, the mean ASR exhibits noticeable fluctuations and occasional collapses, primarily driven by CO3Net. 
These results suggest that excessively strong smoothness constraints can suppress high-frequency perturbation structures that remain important for disrupting texture-dominant palmprint recognition.

Based on the above sweeps, we adopt $(\lambda_{\mathrm{id}}, \lambda_{\mathrm{vis}}, \lambda_{\mathrm{tv}})=(0.2,\,4\times10^{-3},\,2\times10^{-5})$ as the default configuration in subsequent experiments, since these values are near-optimal in their respective sweeps and jointly deliver strong mean ASR while avoiding brittle regimes across the evaluated palmprint-specific models.

\section{Conclusion and Future Work}
\label{sec:conclusion}

We investigated the vulnerability of deep palmprint recognition models to physically realizable adversarial patch attacks under print-and-capture variation. To this end, we proposed \attacker, a capture-aware adversarial patch framework that combines universal patch optimization with a cross-shaped topology, input-conditioned rendering adaptation, stochastic capture-aware synthesis, and auxiliary multi-scale feature guidance. Experiments on Tongji, IITD, and AISEC show that \attacker\ achieves strong attack performance under both untargeted and targeted settings across generic CNN backbones and palmprint-specific recognizers. The proposed method also exhibits favorable cross-model and cross-dataset transferability, while the cross-shaped design improves cross-architecture stability. Although adversarial training can partially reduce the attack success rate, non-negligible residual vulnerability remains, suggesting that generic adversarial hardening alone is insufficient against this class of structured physical perturbations. These findings indicate that deep palmprint recognition models remain vulnerable to structured, capture-aware patch attacks under realistic physical deployment conditions.

\noindent\textbf{Future work} may proceed in several directions. An important extension is broader real-world physical validation under more diverse acquisition conditions. It is also worthwhile to study richer threat models and broader deployment settings, including more flexible patch geometries, multi-patch attacks, and systems that incorporate liveness detection or multimodal authentication. In addition, future work should explore more palmprint-specific defense strategies against structured physical perturbations.

\bibliographystyle{IEEEtran}
\bibliography{ref}

\end{document}